\documentclass[10pt,letterpaper]{article}
\usepackage{envrethink-arxiv}
\usepackage[T1]{fontenc}
\usepackage{lmodern}
\usepackage[numbers,square,comma,sort&compress]{natbib}
\usepackage{amsmath,amssymb}
\usepackage{hyperref}
\usepackage{url}
\usepackage{hyphenat}
\usepackage{xspace}
\usepackage{enumitem}
\usepackage{tabularx}
\usepackage{pdflscape}
\usepackage{longtable}
\usepackage{flafter}
\usepackage{tikz}
\usepackage{graphicx}
\usepackage{booktabs}
\usepackage{multirow}
\usepackage{array}
\usepackage{colortbl}

\definecolor{tblgroup}{RGB}{230,240,252}
\definecolor{tblgain}{RGB}{0,150,70}
\definecolor{tblloss}{RGB}{215,40,40}

\usepackage{placeins}

\usetikzlibrary{arrows.meta,positioning,fit,shapes.geometric,calc}
\usepackage{xspace}
\usepackage{ulem}
\usepackage{xcolor}
\usepackage{fontawesome5}
\usepackage{hyperref}
\newcommand{\ourbench}{SelfImprove-Bench\xspace}

\newcommand{\xbr}[1]{\textcolor{black}{#1}}
\definecolor{darkblue}{RGB}{0,51,102}
\hypersetup{
  colorlinks=true,
  urlcolor=darkblue,
  urlbordercolor=darkblue
}

\newcommand{\repolinks}{%
  \vspace{-2em}
  \begin{center}
    \small
    \setlength{\tabcolsep}{1.2em}%
    \begin{tabular}{@{}c@{}}
      \faGithub\ \href{https://github.com/theseus-labs-rsi/SelfImprove-Bench}%
        {\textcolor{darkblue}{\underline{github.com/theseus-labs-rsi/SelfImprove-Bench}}} \\[1pt]
      \faGlobe\ \href{https://theseus-labs-rsi.github.io/SelfImprove-Bench}%
        {\textcolor{darkblue}{\underline{theseus-labs-rsi.github.io/SelfImprove-Bench}}} \\
    \end{tabular}
  \end{center}
}

\title{SelfImprove-Bench: Evaluating Generalization in Self-Improving Agents}
\newcommand{\shortpapertitle}{SelfImprove-Bench}
\newcommand{\preprintlogos}{%
  \raisebox{-0.5\height}{\includegraphics[width=0.32\linewidth]{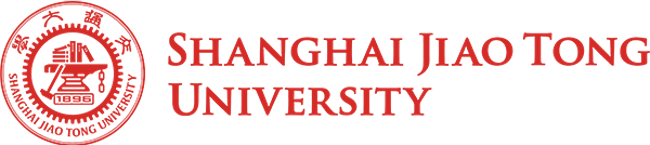}}\hfill
  \raisebox{-0.5\height}{\includegraphics[width=0.32\linewidth]{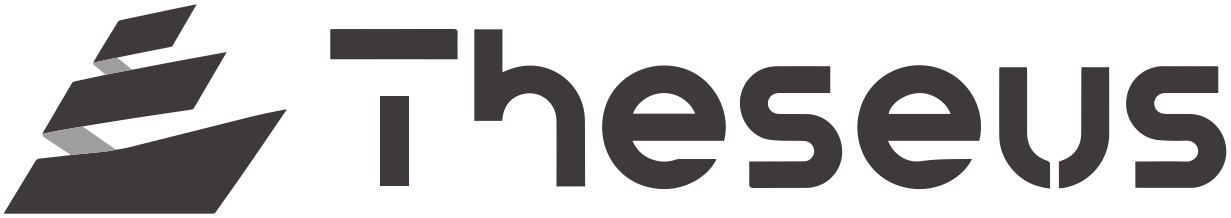}}%
}
\newcommand{\paperauthors}{%
  \mbox{Bangrui Xu\textsuperscript{1,2,3}\thanks{Equal contribution.}}\quad
  \mbox{Jie Yang\textsuperscript{1,2,4}\footnotemark[1]}\quad
  \mbox{Sidi Chen\textsuperscript{1,2,3}\footnotemark[1]}\quad
  \mbox{Xuanhe Zhou\textsuperscript{1,2}\thanks{Corresponding author.}}\quad
  \mbox{Jiantao Qiu\textsuperscript{4}}\\[0.3em]
  \mbox{Qianyu Yang\textsuperscript{3}}\quad
  \mbox{Zijie Wang\textsuperscript{3}}\quad
  \mbox{Fan Wu\textsuperscript{1,2}}\quad
  \mbox{Zhoufutu Wen\textsuperscript{3}\footnotemark[2]}\quad
}
\author{%
  \paperauthors\\[0.5em]
  {\small\textsuperscript{1}Shanghai Jiao Tong University\quad
  \textsuperscript{2}Theseus Labs\quad
  \textsuperscript{3}Humanlaya}\quad
  \textsuperscript{4}Shanghai AI Lab
}
\date{}
\hypersetup{
  hidelinks,
  pdftitle={SelfImprove-Bench: Evaluating Generalization in Self-Improving Agents},
  pdfauthor={Bangrui Xu, Jie Yang, Sidi Chen, Xuanhe Zhou, Jiantao Qiu, Qianyu Yang, Jianpeng Jiao, Zijie Wang, Fan Wu, Zhoufutu Wen}
}

\begin{document}
\maketitle
\repolinks
\begin{abstract}

Recursively self-improving (RSI) systems are increasingly viewed as a promising means of accelerating progress toward artificial general intelligence. They aim to autonomously improve their capabilities by refining their working pipelines and recursively improving the strategies that guide subsequent refinements, with gains generalizing to unseen scenarios. 
However, existing evaluations lack reliable metrics to distinguish capability gains from scenario fitting, sustainable and extensible protocols for unseen tasks and difficulties, and fine-grained attribution of improvement to evolution traces.


To this end, we propose \textbf{\ourbench}, an RSI evaluation framework that organizes tasks into three primary capabilities (i.e., Exploring, Planning, Reasoning) as scenarios for guiding model evolution. We instantiated the tasks on games for controllability and reproducibility, and pairs them with scalable generation and orchestration driven by task templates and difficulty coefficients. Beyond task-level scores, \ourbench organizes evaluation into \textit{generalization ladder levels G0 to G4} that measure how far gains transfer from training conditions. 
Experiments reveal that the evolutionary gains of existing RSI methods are inconsistent across generalization levels. 

\end{abstract}

\begin{figure*}[h]
\centering
\includegraphics[width=.95\linewidth]{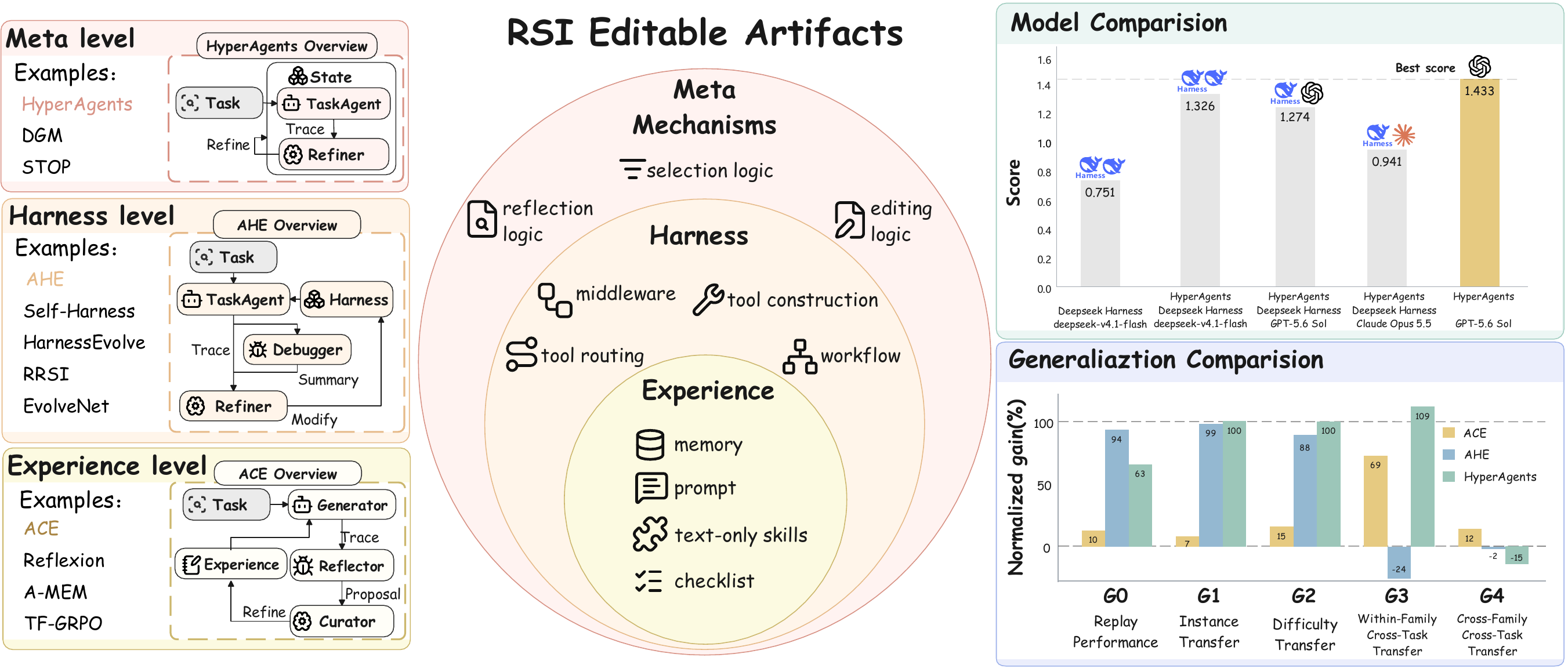}
\vspace{1em}
\caption{\textbf{RSI at a glance.} 
Left and center: representative RSI methods and editable artifacts. 
Top right: performance across different agent configurations. 
Bottom right: capability retention and generalization across G0--G4.}
\label{fig}
\end{figure*}

\section{Introduction}
\label{sec}
Despite rapid advances in large language models (LLMs), conventional development still relies heavily on static datasets and offline training, limiting continuous adaptation to changing environments. Recursive self-improving (RSI) agents address this limitation by iteratively modifying their memories, skills, and harnesses using experience and feedback~\cite{zhang2026hyperagents, zhang2025agentic, lin2026agentic}. A central question is whether these improvements generalize beyond encountered instances to harder tasks, different task types, and ultimately distinct capabilities. Evaluating this progression is essential to understanding RSI as a path toward artificial general intelligence.

Existing benchmarks leave three challenges unresolved (Table~\ref{tab:benchmark_comparison}). \textbf{(i) Systematic generalization assessment.} RSI benchmarks examine selected settings, such as cross-task or cross-difficulty transfer~\cite{zheng2026seagym, huang2026evobench, deng2026finevo}, but lack a unified taxonomy spanning instances, difficulty levels, tasks, and capabilities. \textbf{(ii) Scalable RSI evaluation.} Although some reinforcement learning benchmarks cover broader generalization settings~\cite{stojanovski2025reasoning}, they lack a unified protocol for general-purpose RSI methods. Fixed task pools and limited difficulty ranges further constrain sustained evaluation as agents evolve and approach performance ceilings~\cite{xue2026past, gao2026evoagentbench}. \textbf{(iii) Fine-grained process diagnosis.} Existing analyses rely primarily on aggregate metrics, selected case studies, or persistence ablations~\cite{zhong2026skilllearnbench, zheng2025lifelongagentbench}, leaving the relationship between evolution trajectories and generalization outcomes insufficiently understood.

To address these challenges, we introduce \ourbench, an RSI evaluation framework combining capability-oriented tasks with a generalization ladder. (i) The ladder assesses progressively broader generalization across tasks targeting planning, reasoning, and exploration. It distinguishes five levels: replay performance on the same instances (G0), unseen instances at the same difficulty (G1), new difficulty levels within the same task (G2), different tasks within the same capability category (G3), and tasks across capability categories (G4). (ii) A scalable task generation and orchestration module generates instances using task templates and difficulty coefficients, then assembles sequences with configurable task types, counts, difficulty levels, and replay frequencies. (iii) An agent-driven, fine-grained trace analysis module characterizes refinement semantics, modification scope, capability dimensions, and generalizability to reveal method-specific evolution patterns.

\begin{figure}[h]
    \centering
    \includegraphics[width=\linewidth]{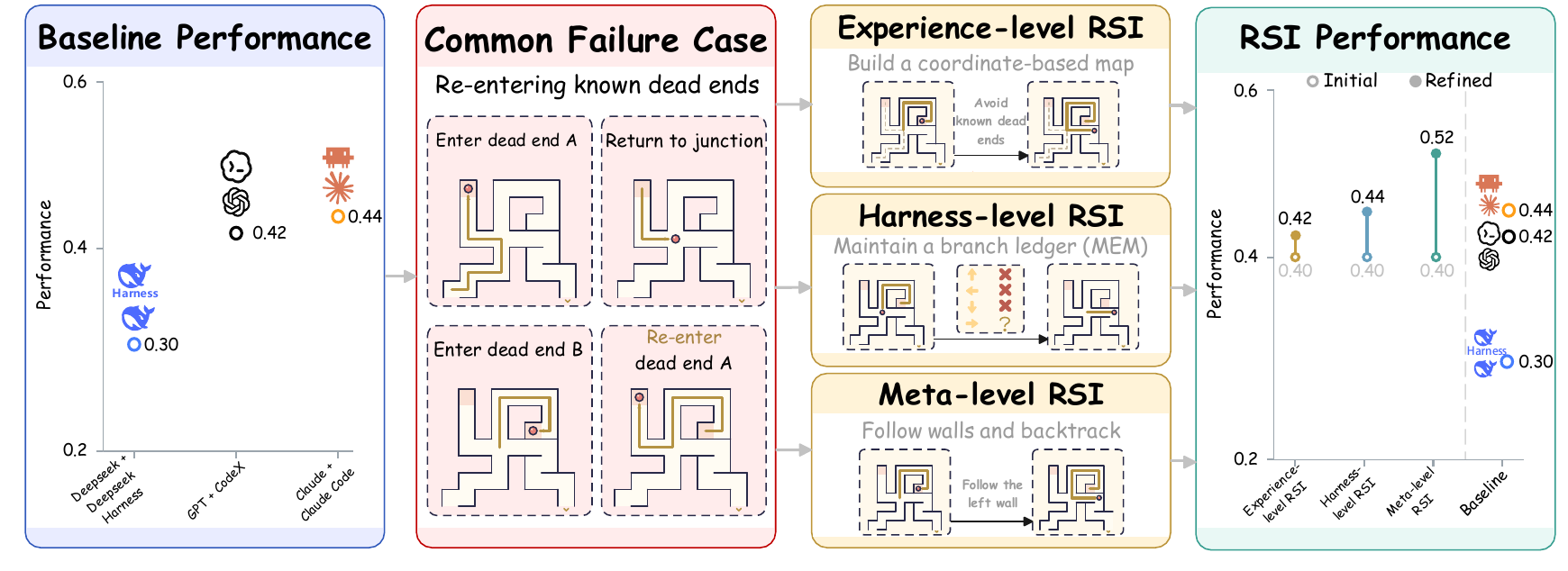}
    \caption{\textbf{Example RSI-driven refinements.}
    The blue panel shows the performance of different agent systems.
    The red panel illustrates a common failure case.
    The yellow panels show refinement examples from different RSI Methods.
    The green panel shows RSI performance before and after refinement}
    \label{fig:refine-example}
\end{figure}

Using \ourbench, we evaluate three representative classes of RSI methods and observe substantial variation across the generalization ladder, including ranking reversals on harder instances and different task types. Note we focus on non-training RSI methods, whose explicit modifications to prompts, memories, skills, and harnesses support a unified evaluation protocol and interpretable evolution traces. Their lower computational requirements also make repeated evaluation more tractable. Experiments across three representative classes of RSI methods reveal substantial differences along the generalization ladder, including ranking reversals on harder instances and different tasks. As illustrated in Figure~\ref{fig}, similar final scores can therefore conceal distinct evolution patterns and transfer ability.

Our main contributions are as follows. 

(1) We introduce a capability-oriented generalization ladder for systematically evaluating RSI across instances, difficulty levels, tasks, and capabilities.

(2) We develop configurable task generation and orchestration to support scalable evaluation across the ladder.

(3) We evaluate representative RSI methods and analyze their evolution traces, revealing differences in improvement effectiveness, generalization, and modification patterns.

\section{Related Work}\label{sec:related}

\begin{table*}[t]
\begingroup
\definecolor{coverageblue}{RGB}{230,240,252}
\definecolor{coveragegreen}{RGB}{0,135,65}
\newcommand{\coveragecheck}{\textcolor{coveragegreen}{$\checkmark$}}
\centering
\small
\setlength{\tabcolsep}{6pt}
\setlength{\arrayrulewidth}{0.3pt}
\renewcommand{\arraystretch}{1.12}
\resizebox{\linewidth}{!}{
\begin{tabular}{l|c|c|cc}
\hline
\rowcolor{coverageblue}
& & & \multicolumn{2}{c}{\textbf{Cross-Task Transfer}} \\
\rowcolor{coverageblue}
\multirow{-2}{*}{\textbf{Work}}
& \multirow{-2}{*}{\textbf{\shortstack{Quantitative Analysis\\of RSI Modifications}}}
& \multirow{-2}{*}{\textbf{\shortstack{Easy-to-Hard\\Transfer}}}
& Within-Capability & Cross-Capability \\
\hline
EvoPathBench~\citep{lin2026beyond} & \coveragecheck & --- & --- & --- \\
EvoAgentBench~\citep{gao2026evoagentbench} & --- & --- & \coveragecheck & --- \\
SEAGym~\citep{zheng2026seagym} & --- & --- & \coveragecheck & \coveragecheck \\
SkillLearnBench~\citep{zhong2026skilllearnbench} & \coveragecheck & --- & --- & --- \\
ServeLearnBench~\citep{zheng2026servelearnbench} & \coveragecheck & --- & --- & --- \\
PAST-Bench~\citep{xue2026past} & \coveragecheck & --- & --- & --- \\
\hline
STOP~\citep{zelikman2024stop} & \coveragecheck & --- & --- & --- \\
HyperAgents~\citep{zhang2026hyperagents} & --- & \coveragecheck & --- & --- \\
ADAS~\citep{hu2025automated} & --- & \coveragecheck & \coveragecheck & \coveragecheck \\
RRSI~\citep{xia2026rrsi} & --- & --- & \coveragecheck & --- \\
\hline
AI4AI-Bench~\citep{chi2026ai4ai} & \coveragecheck & --- & --- & --- \\
RSIBench-Data~\citep{meng2026rsibenchdata} & \coveragecheck & --- & --- & --- \\
Reasoning Gym~\citep{stojanovski2025reasoning} & --- & \coveragecheck & \coveragecheck & \coveragecheck \\
\hline
\ourbench & \coveragecheck & \coveragecheck & \coveragecheck & \coveragecheck \\
\hline
\end{tabular}
}
\caption{\textbf{Comparison of evaluation coverage.} Rows are grouped, from top to bottom, into RSI benchmarks, evaluations accompanying RSI methods, and related evaluations outside RSI. \ourbench is shown separately. Task families are organized by agent ability. For task-type transfer, refinement uses type~$A$, while type~$B$ is held out until evaluation; Same Ability and Cross Ability indicate whether the two types belong to the same ability-defined family. Easy-to-Hard requires evaluation on harder tasks than those used during refinement. Quantitative RSI Trace Analysis requires statistics over refinement trajectories rather than selected examples.}
\label{tab:benchmark_comparison}
\endgroup
\end{table*}

\subsection{RSI Methods} 

We classify RSI methods into three non-exclusive strategies based on their modification targets:

\begin{itemize}
\item \textbf{Experience-level RSI} adapts prompts, memories, and playbooks, as in Reflexion, A-MEM, and ACE \cite{shinn2023reflexion,xu2025amem,zhang2025agentic}.
\item \textbf{Harness-level RSI} modifies tools, workflows, and executable code, as in SICA, Self-Harness, and AHE \cite{robeyns2025self,zhang2026self,lin2026agentic}.
\item \textbf{Meta-level RSI} refines candidate generation or selection, as in DGM's evolving self-modification agent and HyperAgents' editable meta-agent \cite{zhang2026darwin,zhang2026hyperagents}.
\end{itemize}
Other strategies include reinforcement learning (RL), which typically updates policies using reward feedback within a predefined optimization framework \cite{ouyang2022training}.
Data synthesis provides another strategy, seeking to improve model capabilities through generated training data \cite{zelikman2022star}.

RSI methods flexibly integrate and organize these strategies into a continual self-improvement process and has been explored as a potential route toward artificial general intelligence (AGI) \cite{nivel2013bounded}.
RSI evaluation should therefore examine whether improvements achieved through refinement transfer to harder tasks and unseen task types within and across capability categories.

\subsection{RSI Relevant Benchmarks}

\noindent\textbf{RSI-Specific Generalization Evaluation.}
 \label{sec:related-generalization}
Common practices include cross-task testing without explicit capability categories organizing refinement and transfer \cite{zheng2026seagym,wang2026evomembench}, and refinement and evaluation over mixed task streams \cite{huang2026evobench,deng2026finevo,guan2026continualskillbench,zheng2025lifelongagentbench}.
Mixed-stream evaluation examines cumulative performance and experience reuse, but alone does not isolate source-to-target transfer.
Capability-aware protocols often focus on related-task or within-domain generalization \cite{lin2026beyond,gao2026evoagentbench,xia2026rrsi}. 
Capability breakdowns may also serve post-hoc scoring rather than organize refinement and transfer \cite{deng2026finevo}.
Method-specific evaluations additionally demonstrate cross-domain transfer, as in ADAS \cite{hu2025automated}.

Difficulty-based task organization does not necessarily include testing whether gains acquired on easier tasks transfer to harder ones \cite{zheng2026servelearnbench}.
Outside agent RSI, Reasoning Gym provides model-level evaluations of within-domain and cross-domain transfer, alongside difficulty-based evaluations \cite{stojanovski2025reasoning}.
A unified evaluation of agent RSI that combines capability-based refinement, difficulty generalization, and transfer across task categories remains underexplored.

\noindent\textbf{RSI-Specific Trace Level Analysis.}
Process analysis commonly illustrates refinement through selected modification cases \cite{asawa2026continuallearningbench,meng2026rsibenchdata}.
Quantitative approaches assess experience contributions through persistence ablations and memory-free comparisons \cite{xue2026past,wang2026evomembench}, or track capability changes across refinement checkpoints \cite{lin2026beyond}.
Finer analyses can depend on specific artifact formats: SkillLearnBench measures skill quality, usage, and alignment with execution trajectories, requiring explicit skill artifacts \cite{zhong2026skilllearnbench}.
Related work also aggregates semantic categories of code modifications and relates them to performance, as in AI4AI-Bench \cite{chi2026ai4ai}.
However, statistical analysis jointly relating generalization performance, modification types, target capabilities, and the semantic content of individual changes remains underexplored.

\noindent\textbf{Tested Tasks.} Existing RSI evaluations involve tasks like software engineering, terminal interaction, web automation, expert-level reasoning \cite{jimenez2023swe,merrill2026terminal,zhou2023webarena,phan2025humanity,zheng2026seagym},
or construct domain-specific tasks for training-algorithm design and financial workflows \cite{chi2026ai4ai,deng2026finevo}.
However, end-to-end task scores offer limited capability-specific diagnosis,
while fixed task pools may lose discriminative power as methods approach score ceilings.


\section{\ourbench}
\label{sec:bench}

\begin{figure*}[t]
    \centering
    \includegraphics[width=1\linewidth]{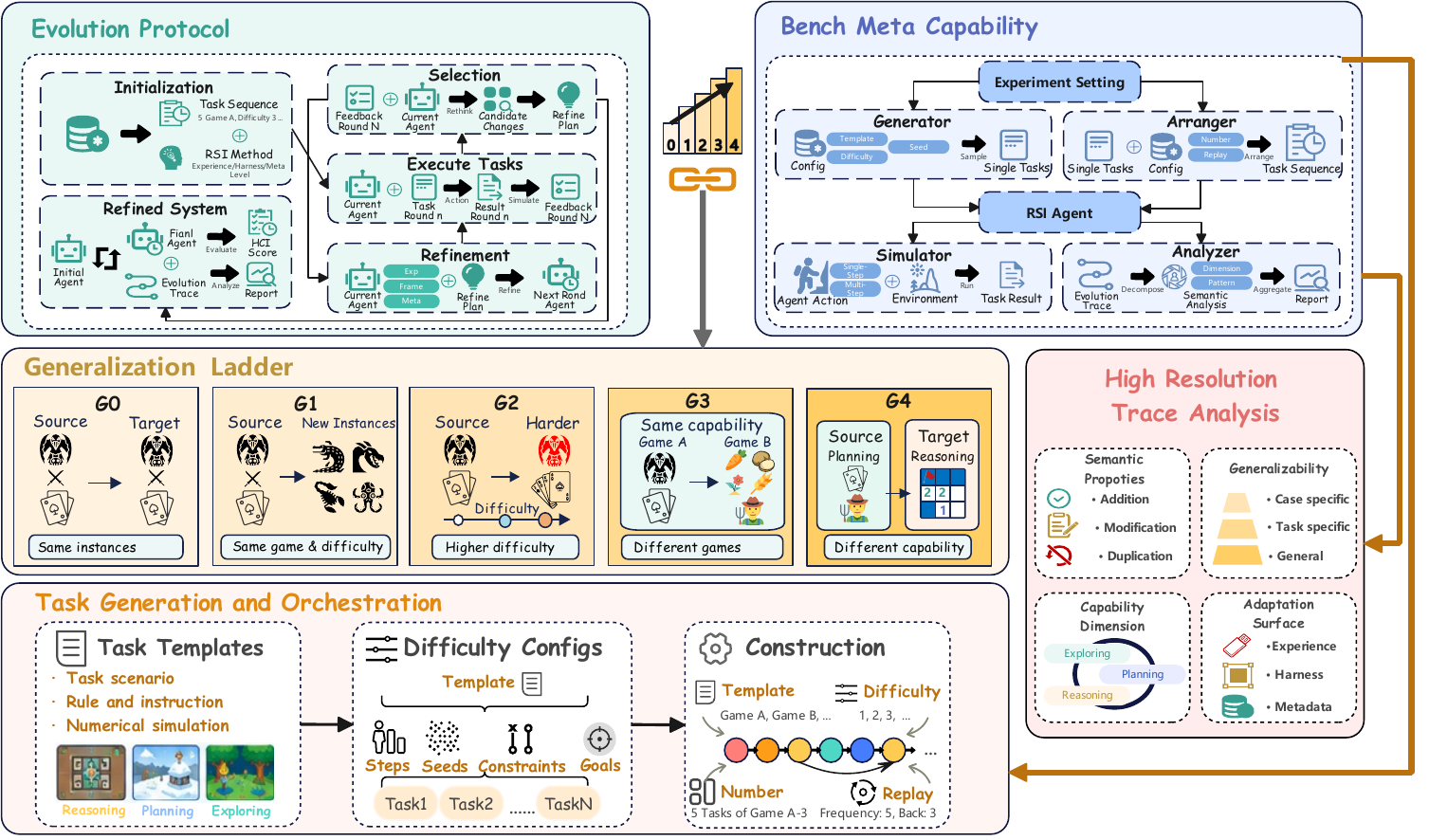}
    \caption{\textbf{\ourbench Overview.} It supports (i) task generation and orchestration targeting primary agentic capabilities to support RSI refinement and (ii) G0--G4 generalization ladder evaluation.} 
    \label{fig:intro}
\end{figure*}

\subsection{Evolution Protocol}

\paragraph{Definition.} Following the definition of RSI in ~\cite{duan2026aibuilthumansgenuine}, we call a refinement gain self-improvement to the extent that it persists on instances, difficulties, and tasks never exposed during refinement, and task fitting to the extent that it is confined to the training instances.

\paragraph{Task construction and orchestration.}
Each instance $\tau=G_g(d,\xi)$ is generated from game $g$, difficulty $d$ , and seed $\xi$.
The configuration $\theta$ first specify the games to include, then assigns difficulty levels and random seeds to generate their instances.
It specifies how many instances enter the held-in and held-out sequences and in       order, yielding $\mathcal{S}_{\mathrm{in}}=(\tau_1,\ldots,\tau_N)$ for execution and refinement and $\mathcal{S}_{\mathrm{out}}=(\tilde{\tau}_1,\ldots,\tilde{\tau}_M)$ for evaluation.
Replay settings determine which held-in tasks can recur, how frequently, and how far back they are selected, counting only non-replay tasks.
Replays do not trigger refinement. $N$ includes replays.
Detailed insertion rules are provided in Appendix~\ref{sec:replay-protocol}.
The corresponding task collections $\mathcal{D}_{\mathrm{in}}$ and $\mathcal{D}_{\mathrm{out}}$ omit order and replay multiplicity.
The generator constructs the held-out pool first and rejects duplicate task-descriptor hashes, excluding \texttt{task\_id}, when generating non-replay held-in tasks (Appendix~\ref{sec:task-configurations}).
Game categorization is described in Section~\ref{sec:task-categorization}. Task configurations are provided in Appendix~\ref{sec:task-configurations}.

We instantiate the framework on games, which satisfy the controllability and reproducibility requirements of RSI evaluation. Our game rules are explicit and state spaces are bounded, making it feasible to map planning, reasoning, and exploring onto different games and avoid the confounding of multiple abilities that complicates real-world tasks. Games inherently embody core characteristics of real-world problems such as partial observability, resource constraints, and multi-step decision making. Separating self-improvement from task fitting requires fresh instances and controlled difficulty, which games most readily provide. We use them as a controlled testbed that complements, rather than replaces, real-world RSI benchmarks. In addition, these games are agent tasks mainly based on text interaction and are related to the NLP community.

\paragraph{Agent initialization.}
For each RSI method, the initial task agent is $A_0=(B,H_0)$, comprising a common backbone model $B$ and a method-specific initial harness $H_0$.
Backbone settings are reported in Section~\ref{sec:experimental-setup}.

\paragraph{Refinement and selection.}
At step $t$, the current agent $A_{t-1}=(B,H_{t-1})$ executes $\tau_t\in\mathcal{S}_{\mathrm{in}}$ through the game's action interface, producing an interaction trace and task score.
Each task execution, except replays, is followed by refinement using benchmark-provided feedback.
During refinement, the method may use tools to inspect and modify its designated workspace, but may not access \ourbench modules or directly access, execute, or evaluate task instances.
Feedback is supplied only through the prescribed interface (Appendix~\ref{sec:feedback-specification}).
When the method proposes candidate harnesses, \ourbench evaluates their agents on the same task instance $\tau_t$ and supplies the results, together with the current agent's result, to the method's selection procedure.
Selection follows the method-specific rule and may retain the current harness. The resulting agent $A_t=(B,H_t)$ proceeds to the next task.
Candidate validation uses held-in tasks only, without accessing $\mathcal{S}_{\mathrm{out}}$.
The backbone $B$ remains fixed, and refinement ends when $\mathcal{S}_{\mathrm{in}}$ is completed.

\paragraph{Evaluation and analysis.}
Evaluation uses saved system states and task results according to the protocols in Section~\ref{sec:metrics}.
Process analysis extracts individual modification proposals from saved refinement reports and edit records, then annotates and aggregates them using the criteria in Section~\ref{sec:process_analysis}. Annotation details are provided in Appendix~\ref{sec:annotation-protocol}.

\subsection{Task Categorization}
\label{sec:task-categorization}
For rule-constrained tasks with fixed action interfaces, we distinguish three interacting capabilities: \textbf{Exploration}, acquiring task-relevant information through observation and interaction. 
\textbf{Reasoning}, interpreting rules, deriving implications, and performing computations.
\textbf{Planning}, organizing action sequences to achieve goals under constraints.
Games are categorized by their primary capability demands, providing context for analyzing modification targets.

\subsection{Generalization Ladder}
\label{sec:metrics}

The generalization ladder evaluates refinement gain, agent transfer, and refinement-mechanism effectiveness across G0--G4.

\noindent\textbf{G0: Replay performance.} For replay event $j$, let $A_{\mathrm{init},j}$ denote the agent before its first execution of the replayed task, and $A_{\mathrm{replay},j}$ the agent executing the current replay.
At G0, compare $A_{\mathrm{replay},j}$ with $A_{\mathrm{init},j}$ on the replayed task $\tau_j$.

\noindent\textbf{G1--G4: Agent generalization.} Relative to a source game $g$ at difficulty $d_0$, $\mathcal{D}_1$ contains instances $G_g(d_0,\xi')$ and $\mathcal{D}_2$ contains $G_g(d_1,\xi')$ with $d_1>d_0$.
Collections $\mathcal{D}_1$ and $\mathcal{D}_2$ are drawn from $\mathcal{D}_{\mathrm{out}}$, whose task descriptors are kept distinct from non-replay held-in descriptors by the construction protocol.
Collections $\mathcal{D}_3$ and $\mathcal{D}_4$ contain other games $g'$ not used for refinement, with $c(g')=c(g)$ and $c(g')\ne c(g)$, respectively. $c(g)$ denotes the primary capability category.
For mixed-source sequences, source games and capability categories are specified for each transfer comparison.
At G1--G4, compare $A_T$ with $A_0$ on the same target instances.

\paragraph{Metrics.} Across G0--G4, we use the HCI (Headroom-Closed Index)~\cite{duan2026aibuilthumansgenuine} to normalize gains relative to the initial gap to a reference score. We first average the baseline and comparison scores over target instances, then compute HCI from the resulting means. Let $s_b$ and $s_a$ denote these mean baseline and comparison scores and $s_r$ the reference score :
\begin{equation}
\mathrm{HCI}=\frac{s_a-s_b}{s_r-s_b}.
\end{equation}

\subsection{Semantic Process Analysis}
\label{sec:process_analysis}


\paragraph{Semantic relations.}
These indicate whether a proposal introduces new content,
contradicts or revises existing proposals, 
or duplicates them,
relative to the parent proposal snapshot and earlier proposals in the same stage.

\paragraph{Applicability scope.}

This distinguishes task-specific proposals, 
proposals applicable across tasks sharing a capability, 
and fully general proposals.

\paragraph{Functional targets.}
These identify which capability a proposal targets:
exploration, reasoning, planning, or execution.
Execution concerns operational mechanisms such as tool invocation,
output formatting, retries, and error recovery.

\paragraph{Modification surfaces.}
These identify whether the change
affects contextual artifacts, the execution harness,
or the refinement mechanism itself.

\begin{table*}[!t]
\centering
\begingroup
\definecolor{laddergroup}{RGB}{230,240,252}
\definecolor{laddergain}{RGB}{0,135,65}
\definecolor{ladderloss}{RGB}{200,35,35}
\newcommand{\ladderpos}[1]{\textcolor{laddergain}{#1}}
\newcommand{\ladderneg}[1]{\textcolor{ladderloss}{#1}}
\fontsize{7}{8.5}\selectfont
\setlength{\tabcolsep}{2pt}
\setlength{\arrayrulewidth}{0.3pt}
\renewcommand{\arraystretch}{1.18}
\resizebox{\textwidth}{!}{%
\begin{tabular}{ll|rrr|rrr|rrr|rrr}
\hline
\rowcolor{laddergroup}
 & & \multicolumn{3}{c|}{\textbf{G0}} & \multicolumn{3}{c|}{\textbf{G1}} & \multicolumn{3}{c|}{\textbf{G2}} & \multicolumn{3}{c}{\textbf{G3}} \\
\rowcolor{laddergroup}
Task & Method & Init. & Final & \textbf{HCI (\%)} & Init. & Final & \textbf{HCI (\%)} & Init. & Final & \textbf{HCI (\%)} & Init. & Final & \textbf{HCI (\%)} \\
\hline
\textbf{Farm} & ACE & 1.078 & 1.067 & \ladderneg{-10.2} & 0.997 & 1.145 & \ladderpos{+81.3} & 0.784 & 0.859 & \ladderpos{+14.6} & 0.533 & 0.596 & \ladderpos{+7.0} \\
 & AHE & 0.768 & 1.009 & \ladderpos{\textbf{+58.7}} & 1.105 & 1.080 & \ladderneg{-33.8} & 1.073 & 1.069 & \ladderneg{-1.8} & 0.674 & 0.792 & \ladderpos{\textbf{+15.5}} \\
 & Hyper & 1.046 & 1.084 & \ladderpos{+28.0} & 1.048 & 1.179 & \ladderpos{\textbf{+100.0}} & 0.857 & 0.942 & \ladderpos{\textbf{+19.3}} & 0.597 & 0.596 & \ladderneg{-0.1} \\
\hline
\textbf{Combat} & ACE & 0.855 & 0.911 & \ladderpos{+9.7} & 0.533 & 0.592 & \ladderpos{+6.6} & 0.227 & 0.385 & \ladderpos{+14.5} & 0.997 & 1.123 & \ladderpos{+69.2} \\
 & AHE & 1.065 & 1.410 & \ladderpos{\textbf{+93.8}} & 0.674 & 1.427 & \ladderpos{+99.2} & 0.301 & 1.196 & \ladderpos{+88.4} & 1.105 & 1.087 & \ladderneg{-24.3} \\
 & Hyper & 0.881 & 1.226 & \ladderpos{+62.5} & 0.597 & 1.433 & \ladderpos{\textbf{+100.0}} & 0.233 & 1.314 & \ladderpos{\textbf{+100.0}} & 1.048 & 1.191 & \ladderpos{\textbf{+109.2}} \\
\hline
\textbf{Sweepmine} & ACE & 0.239 & 0.193 & \ladderneg{-6.0} & 0.134 & 0.316 & \ladderpos{+21.0} & 0.043 & 0.234 & \ladderpos{+20.0} & 0.094 & 0.067 & \ladderneg{-3.0} \\
 & AHE & 0.432 & 0.420 & \ladderneg{-2.0} & 0.150 & 0.436 & \ladderpos{\textbf{+33.7}} & 0.107 & 0.350 & \ladderpos{\textbf{+27.2}} & 0.097 & 0.077 & \ladderneg{-2.2} \\
 & Hyper & 0.193 & 0.216 & \ladderpos{\textbf{+2.8}} & 0.086 & 0.205 & \ladderpos{+12.9} & 0.014 & 0.117 & \ladderpos{+10.5} & 0.100 & 0.085 & \ladderneg{\textbf{-1.7}} \\
\hline
\textbf{Hero} & ACE & 0.169 & 0.124 & \ladderneg{-5.4} & 0.094 & 0.154 & \ladderpos{+6.7} & 0.017 & 0.124 & \ladderpos{+10.9} & 0.134 & 0.234 & \ladderpos{+11.5} \\
 & AHE & 0.187 & 0.244 & \ladderpos{\textbf{+7.0}} & 0.097 & 0.526 & \ladderpos{\textbf{+47.5}} & 0.054 & 0.257 & \ladderpos{\textbf{+21.4}} & 0.150 & 0.318 & \ladderpos{\textbf{+19.8}} \\
 & Hyper & 0.087 & 0.070 & \ladderneg{-1.9} & 0.100 & 0.085 & \ladderneg{-1.7} & 0.026 & 0.082 & \ladderpos{+5.7} & 0.086 & 0.264 & \ladderpos{+19.4} \\
\hline
\textbf{MC} & ACE & 0.162 & 0.077 & \ladderneg{-10.1} & 0.000 & 0.020 & \ladderpos{+2.0} & 0.000 & 0.350 & \ladderpos{+35.0} & 0.400 & 0.400 & \textbf{+0.0} \\
 & AHE & 0.903 & 0.728 & \ladderneg{-179.4} & 0.505 & 0.720 & \ladderpos{\textbf{+43.4}} & 1.000 & 0.600 & \ladderneg{N/A} & 0.400 & 0.400 & \textbf{+0.0} \\
 & Hyper & 0.278 & 0.483 & \ladderpos{\textbf{+28.4}} & 0.050 & 0.220 & \ladderpos{+17.9} & 0.000 & 0.400 & \ladderpos{\textbf{+40.0}} & 0.400 & 0.200 & \ladderneg{-33.3} \\
\hline
\textbf{Astray} & ACE & 0.318 & 0.395 & \ladderpos{+11.4} & 0.400 & 0.420 & \ladderpos{+3.3} & 0.222 & 0.211 & \ladderneg{-1.4} & 0.000 & 0.050 & \ladderpos{+5.0} \\
 & AHE & 0.617 & 0.769 & \ladderpos{\textbf{+39.6}} & 0.400 & 0.440 & \ladderpos{+6.7} & 0.222 & 0.378 & \ladderpos{\textbf{+20.0}} & 0.505 & 0.550 & \ladderpos{\textbf{+9.1}} \\
 & Hyper & 0.274 & 0.265 & \ladderneg{-1.2} & 0.400 & 0.520 & \ladderpos{\textbf{+20.0}} & 0.244 & 0.222 & \ladderneg{-2.9} & 0.050 & 0.000 & \ladderneg{-5.3} \\
\hline
\end{tabular}%
}
\endgroup

\caption{\textbf{Performance across G0--G3.} Bold marks the highest available HCI for each game at each level, including ties. For G3, each row is the source game and Final is its agent's score on the paired target (Farm--Combat, Sweepmine--Hero, MC--Astray).}
\label{tab:adaptivity-all}
\end{table*}

\section{Experiments}
\label{sec:experiments}

We examine the gains from iterative refinement in ACE, AHE, and HyperAgents and whether these gains extend to unseen instances, harder tasks, and other games. We further analyze their refinement trajectories, task scheduling, and execution-harness refinement.

\subsection{Experimental Setup}
\label{sec:experimental-setup}

\paragraph{Tested Baselines.}
We evaluate ACE, AHE, and HyperAgents as representative methods of experience-level, harness-level, and meta-level RSI, respectively (Section~\ref{sec:related}). ACE maintains an evolving structured playbook; AHE revises the prompt, tools, and execution harness based on task feedback; and HyperAgents uses an improver to modify the task agent's code repository. We vary the backbone model, including GPT-5.6 Sol and Claude Opus 5.5, and the initial execution harness, including Codex and DeepSeek Harness, to assess how these choices affect RSI performance.


\paragraph{Configuration.} Farm, Combat, Sweepmine, and Hero support difficulty levels 1--5. For MC and Astray, evaluation uses five fixed tasks per game from the official GameWorld~\citep{ouyang2026gameworld} evaluation sets.

\subsection{Main Results}
\label{sec:main-results}

\begin{figure*}[!tb]
\centering
\vspace{1em}
\includegraphics[width=0.92\textwidth]{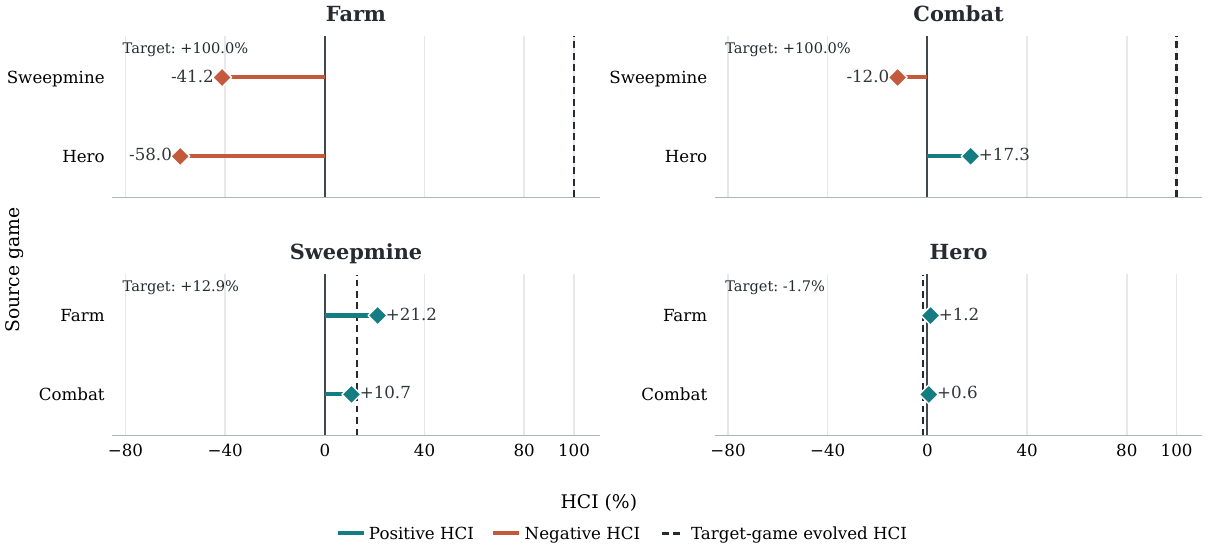}
\caption{\textbf{HyperAgents cross-category transfer (G4).} Each panel is a target game; lines give the HCI of the agent refined on each source game. Dashed lines mark the HCI of refinement directly on the target.}
\label{fig:transfer-hci}
\vspace{1em}
\end{figure*}

We present the HCI scores of each RSI method at different levels (G0 to G4) in Tablee~\ref{tab:adaptivity-all} and Figure~\ref{fig:transfer-hci}.

\noindent\textbf{(G0) Retention on Previously Encountered Instances.} Refinement does not always preserve performance on previously encountered instances: replay scores fall below historical scores in 8 of the 18 method--task pairs. AHE on MC and ACE on Sweepmine improve on G1 but decline on G0.

\noindent\textbf{(G1) Gains on Held-Out Instances.}
\label{sec:overall-performance}
Most methods improve on unseen instances at the training difficulty: 16 of the 18 method--task pairs have positive HCI. HyperAgents has the highest HCI on Farm, Combat, and Astray, and AHE on Sweepmine, Hero, and MC. 
A stronger initial agent does not necessarily stay ahead. On Farm, AHE starts above HyperAgents but finishes below it, with HCI of $-33.8\%$ and $100.0\%$.

\noindent\textbf{(G2) Generalization to Harder Instances.}
\label{sec:main-generalization}
Gains at the training difficulty do not necessarily extend to harder instances. ACE on Astray improves on G1 but declines on G2, and rankings can reverse: on Farm, HyperAgents beats AHE at the training difficulty but trails it on the Hard held-out set. On MC's Hard group, AHE falls from 1.000 to 0.600.

\noindent\textbf{(G3) Within-Category Transfer.}
Within-category transfer can help or hurt, depending on direction. Among the Reasoning games, all three methods improve from Hero to Sweepmine but degrade in the reverse direction.
 An agent refined on another game of the same category can even outperform direct refinement on the target: HyperAgents reaches 109.2\% HCI from Combat to Farm, above the 100.0\% from refinement on Farm itself.

\begin{table*}[!tb]
\centering
\begingroup
\definecolor{tracegroup}{RGB}{230,240,252}
\fontsize{6}{7.5}\selectfont
\setlength{\tabcolsep}{1pt}
\setlength{\arrayrulewidth}{0.3pt}
\renewcommand{\arraystretch}{1.18}
\resizebox{\textwidth}{!}{%
\begin{tabular}{ll|rrrr|rrrr|rrr|rrr}
\hline
\rowcolor{tracegroup}
 & & \multicolumn{4}{c|}{\textbf{Records}} & \multicolumn{4}{c|}{\textbf{Functional target (\%)}} & \multicolumn{3}{c|}{\textbf{Applicability scope (\%)}} & \multicolumn{3}{c}{\textbf{Modification surface (\%)}} \\
\rowcolor{tracegroup}
Method & Game & All & Rev. (\%) & Dup. (\%) & $N$ & Explor. & Reason. & Plan. & Exec. & Task & Category & General & Prompt/Mem. & Harness & Meta \\
\hline
\textbf{ACE} & Farm & 28 & 0.00 & 10.71 & 25 & 0.0 & 84.0 & 48.0 & 4.0 & 96.0 & 4.0 & 0.0 & 100.0 & 0.0 & 0.0 \\
 & Combat & 51 & 3.92 & 17.65 & 40 & 22.5 & 95.0 & 52.5 & 7.5 & 100.0 & 0.0 & 0.0 & 100.0 & 0.0 & 0.0 \\
 & Hero & 78 & 26.92 & 12.82 & 46 & 0.0 & 82.6 & 67.4 & 2.2 & 32.6 & 65.2 & 2.2 & 100.0 & 0.0 & 0.0 \\
 & Sweepmine & 44 & 11.36 & 6.82 & 36 & 8.3 & 72.2 & 41.7 & 8.3 & 100.0 & 0.0 & 0.0 & 100.0 & 0.0 & 0.0 \\
 & MC & 65 & 16.92 & 13.85 & 43 & 18.6 & 53.5 & 53.5 & 25.6 & 100.0 & 0.0 & 0.0 & 100.0 & 0.0 & 0.0 \\
 & Astray & 66 & 12.12 & 12.12 & 35 & 17.1 & 65.7 & 25.7 & 11.4 & 100.0 & 0.0 & 0.0 & 100.0 & 0.0 & 0.0 \\
\hline
\textbf{AHE} & Farm & 91 & 32.97 & 13.19 & 50 & 8.0 & 32.0 & 8.0 & 68.0 & 76.0 & 18.0 & 6.0 & 38.0 & 62.0 & 0.0 \\
 & Combat & 56 & 26.79 & 14.29 & 23 & 0.0 & 43.5 & 52.2 & 30.4 & 91.3 & 0.0 & 8.7 & 39.1 & 60.9 & 0.0 \\
 & Hero & 397 & 21.16 & 46.10 & 89 & 16.9 & 82.0 & 27.0 & 25.8 & 91.0 & 4.5 & 4.5 & 98.9 & 1.1 & 0.0 \\
 & Sweepmine & 99 & 27.27 & 18.18 & 53 & 5.7 & 50.9 & 41.5 & 26.4 & 100.0 & 0.0 & 0.0 & 71.7 & 28.3 & 0.0 \\
 & MC & 210 & 40.48 & 15.71 & 82 & 17.1 & 24.4 & 82.9 & 4.9 & 95.1 & 2.4 & 2.4 & 98.8 & 1.2 & 0.0 \\
 & Astray & 215 & 40.00 & 18.60 & 48 & 14.6 & 47.9 & 60.4 & 12.5 & 97.9 & 2.1 & 0.0 & 100.0 & 0.0 & 0.0 \\
\hline
\textbf{Hyper} & Farm & 434 & 55.53 & 17.05 & 102 & 2.9 & 77.5 & 56.9 & 12.7 & 42.2 & 46.1 & 11.8 & 96.1 & 3.9 & 0.0 \\
 & Combat & 265 & 36.60 & 30.94 & 80 & 10.0 & 62.5 & 43.8 & 41.2 & 66.2 & 31.2 & 2.5 & 46.2 & 53.8 & 0.0 \\
 & Hero & 383 & 18.80 & 67.62 & 47 & 0.0 & 57.4 & 42.6 & 6.4 & 80.9 & 14.9 & 4.3 & 100.0 & 0.0 & 0.0 \\
 & Sweepmine & 386 & 38.08 & 43.52 & 57 & 14.0 & 64.9 & 50.9 & 8.8 & 59.6 & 29.8 & 10.5 & 100.0 & 0.0 & 0.0 \\
 & MC & 141 & 40.43 & 36.88 & 22 & 22.7 & 54.5 & 81.8 & 45.5 & 81.8 & 9.1 & 9.1 & 95.5 & 4.5 & 0.0 \\
 & Astray & 276 & 51.81 & 18.48 & 37 & 21.6 & 45.9 & 64.9 & 2.7 & 89.2 & 10.8 & 0.0 & 97.3 & 2.7 & 0.0 \\
\hline
\end{tabular}%
}
\endgroup

\caption{\textbf{Improvement statistics across six games.} Recorded and retained opinions are compared across ACE, AHE, and HyperAgents by functional target, applicability scope, and modification surface. All/$N$: recorded/retained counts; Rev./Dup.: revision/duplicate rates (\%).}
\label{tab:improvement-path}
\end{table*}

\noindent\textbf{(G4) Cross-Category Transfer.}
\label{sec:generalization}
Some HyperAgents improvements remain effective across capability categories, but not consistently: five of the eight directions in Figure~\ref{fig:transfer-hci} have positive HCI. The Farm-refined agent on Sweepmine and Hero, and the Combat-refined agent on Hero, exceed agents refined directly on those targets.

Overall, the evaluated methods show inconsistent refinement gains across generalization levels: improvements on unseen instances do not necessarily extend to higher difficulties or other tasks, and some settings exhibit performance degradation.

\subsection{RSI Trajectory Analysis}
\label{sec:improvement-mechanisms}

The generalization ladder shows where gains persist; trajectory analysis examines what the methods change. We use GPT-5.6 Sol to split improvement artifacts into atomic opinions and annotate their semantic relations, applicability scope, functional targets, and modification surfaces. Table~\ref{tab:improvement-path} covers the 915 opinions retained in the final snapshots of the six games.

Task-specific refinement can coexist with transferable gains.
Task-specific opinions predominate in all three methods; category-general or general ones make up 14.2\% of ACE's retained opinions, 7.8\% of AHE's, and 36.5\% of HyperAgents'. Yet G3 and G4 show positive transfer across games, sometimes exceeding direct refinement on the target.

Refinement emphasizes solving observed problems over exploring alternatives.
Reasoning and planning appear in 61.2\% and 49.6\% of retained opinions, versus 11.0\% for exploration. Current RSI methods may rely on observed trajectories to find and fix problems rather than explore alternatives.

Main-experiment changes concentrate on the task agent.
87.9\% of retained opinions target prompts and memory and 12.1\% the execution harness; none modify the improver or the improvement process.

\begin{figure*}[!t]
\centering
\vspace{1em}
\includegraphics[width=\textwidth]{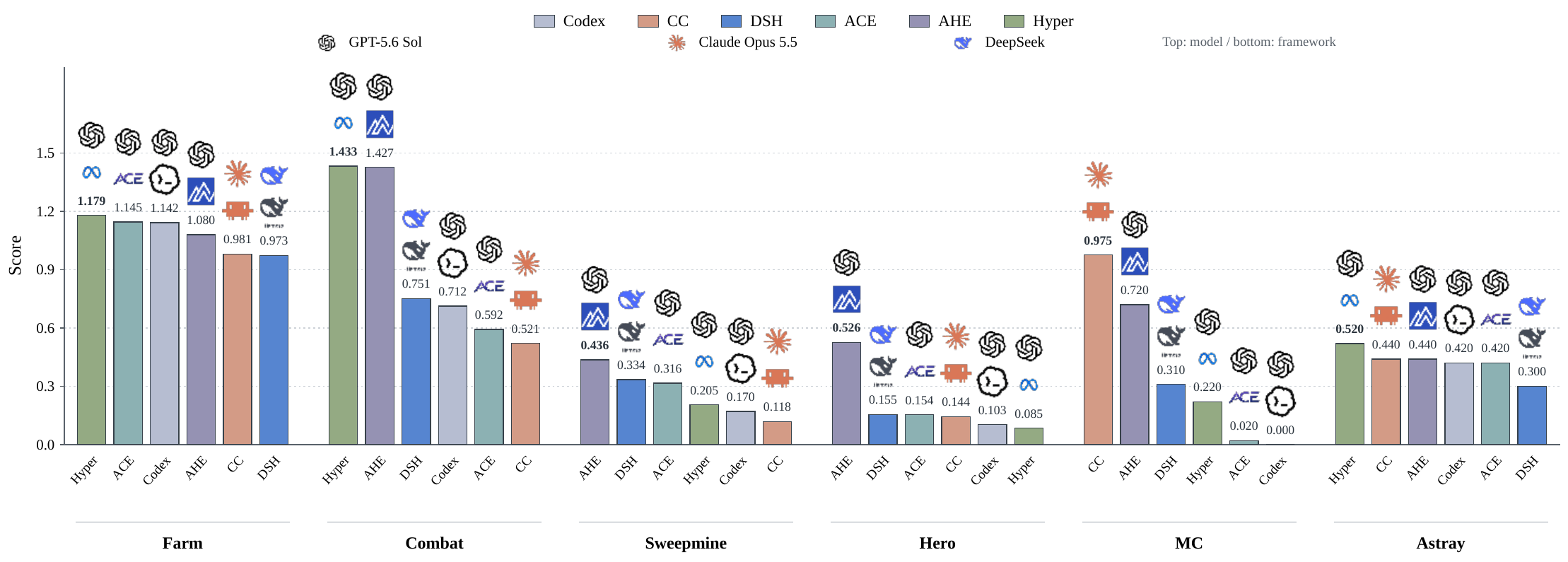}
\caption{\textbf{Comparison with general-purpose agents.} Medium held-out scores of general-purpose agents and refined agents, ranked within each game.}
\label{fig:agent-comparison}
\end{figure*}

More active revision may contribute to better refinement outcomes.
ACE, AHE, and HyperAgents revise 14.2\%, 30.6\%, and 40.2\% of their recorded opinions, respectively. AHE and HyperAgents revise existing opinions more frequently and generally achieve better final performance than ACE, suggesting that effective refinement may depend not only on adding new opinions but also on continually revising existing strategies.

Frequent revision does not eliminate duplicate opinions.
HyperAgents has the highest revision rate, yet on Hero its duplicate rate reaches 67.6\%, against 46.1\% for AHE and 12.8\% for ACE.

\subsection{Ablation and Extended Analysis}
\label{sec:ablation}

\subsubsection{Comparison with General-Purpose Agents}
\label{sec:agent-comparison}

Figure~\ref{fig:agent-comparison} compares the refined agents with Codex, Claude Code, and DeepSeek Harness on Medium held-out tasks. Refined agents score highest on five of the six games: HyperAgents on Farm, Combat, and Astray, and AHE on Sweepmine and Hero. On MC, Claude Code outperforms all refined agents.

\subsubsection{Effect of Task Scheduling}
\label{sec:learning-dynamics}

We evaluate four training task schedules using HyperAgents.

Random medium-difficulty training outperforms fixed-instance training on both Combat held-out sets and on Hard Hero tasks (Table~\ref{tab:curriculum}): fixed-instance training scores 0.772 and 0.507 on Combat and 0.100 and 0.017 on Hero, against 1.433 and 1.314, and 0.085 and 0.082, for random medium-difficulty training. On Medium Hero tasks, however, fixed-instance training scores higher (0.100 versus 0.085).

\begin{table}[!htbp]
\centering
\begingroup
\definecolor{scheduleblue}{RGB}{230,240,252}
\small
\setlength{\tabcolsep}{4pt}
\setlength{\arrayrulewidth}{0.3pt}
\renewcommand{\arraystretch}{1.05}
\begin{tabularx}{.8\textwidth}{l|*{2}{>{\raggedleft\arraybackslash}X}|*{2}{>{\raggedleft\arraybackslash}X}}
\hline
\rowcolor{scheduleblue}
 & \multicolumn{2}{c|}{\textbf{Combat}} & \multicolumn{2}{c}{\textbf{Hero}} \\
\rowcolor{scheduleblue}
Schedule & Medium & Hard & Medium & Hard \\
\hline
Random medium         & \textbf{1.433} & \textbf{1.314} & 0.085 & \textbf{0.082} \\
Fixed medium instance & 0.772 & 0.507 & 0.100 & 0.017 \\
Mixed difficulty      & 1.388 & 1.195 & 0.145 & 0.037 \\
Increasing difficulty & 1.332 & 1.080 & \textbf{0.152} & 0.024 \\
\hline
\end{tabularx}
\endgroup
\caption{\textbf{Effect of training task scheduling.} Medium and Hard held-out scores on Combat and Hero under random medium-difficulty, fixed-instance, mixed-difficulty, and increasing-difficulty training. Bold marks the highest score in each column.}
\label{tab:curriculum}
\end{table}


Harder training instances do not necessarily help. Mixed and increasing difficulty score 1.195 and 1.080 on Hard Combat tasks, below the 1.314 of random medium-difficulty training, and 0.037 and 0.024 on Hard Hero tasks, below 0.082. Both also score lower on Medium Combat tasks, but higher on Medium Hero tasks (0.145 and 0.152 versus 0.085), indicating that their benefits depend on the game and evaluation difficulty.

\subsubsection{Exploratory Experiment: Directly Refining an Execution Harness}
\label{sec:harness-feasibility}

\begin{table}[!htbp]
\centering
\begingroup
\definecolor{refinementblue}{RGB}{230,240,252}
\small
\setlength{\arrayrulewidth}{0.3pt}
\renewcommand{\arraystretch}{1.1}
\begin{tabularx}{0.6\textwidth}{l|>{\raggedright\arraybackslash}X|r}
\hline
\rowcolor{refinementblue}
Task & Configuration & Score $\uparrow$ \\
\hline
Combat & DeepSeek Harness                 & 0.751 \\[2pt]
       & HyperDSH (GPT-5.6 Sol)            & 1.274 \\
       & HyperDSH (Claude Opus 5.5)        & 0.941 \\
       & HyperDSH (DeepSeek)               & 1.326 \\
       & HyperDSH (DeepSeek, fixed improver) & 1.207 \\[2pt]
       & HyperAgents                       & \textbf{1.433} \\
\hline
\end{tabularx}
\endgroup
\caption{\textbf{Direct execution-harness refinement.} Medium held-out Combat scores of the original DeepSeek Harness, HyperDSH variants, and the main-experiment HyperAgents agent. HyperDSH is a variant that applies HyperAgent's improvement mechanisms to Deepseek Harness.}
\label{tab:hyperdsh-feasibility}
\end{table}
We evaluate HyperDSH's direct refinement of DeepSeek Harness on Combat, comparing it with the original DeepSeek Harness and the HyperAgents agent refined in the main experiment (Table~\ref{tab:hyperdsh-feasibility}).
All three HyperDSH variants outperform DeepSeek Harness on Medium held-out tasks; the DeepSeek variant performs best among them at 1.326, below HyperAgents at 1.433.
HyperDSH (DeepSeek, fixed improver), which keeps the improver unchanged while continuing to refine DeepSeek Harness, scores 1.207 on Medium, compared with 1.326 for HyperDSH (DeepSeek).

\begin{table}[!htbp]
\centering
\begingroup
\definecolor{refinementblue}{RGB}{230,240,252}
\small
\setlength{\arrayrulewidth}{0.3pt}
\renewcommand{\arraystretch}{1.1}
\begin{tabularx}{0.95\linewidth}{l|>{\raggedright\arraybackslash}X|rr|rrr}
\hline
\rowcolor{refinementblue}
Task & Configuration & All & $N$ & Prompt/Mem. (\%) & Harness (\%) & Meta (\%) \\
\hline
Combat & DeepSeek Harness                 & -- & -- & -- & -- & -- \\[2pt]
       & HyperDSH (GPT-5.6 Sol)            & 107 & 44 & 79.5 & 0.0 & 20.5 \\
       & HyperDSH (Claude Opus 5.5)        & 57 & 31 & 54.8 & 0.0 & 45.2 \\
       & HyperDSH (DeepSeek)               & 127 & 37 & 59.5 & 0.0 & 40.5 \\
       & HyperDSH (DeepSeek, fixed improver) & 85 & 27 & 96.3 & 3.7 & 0.0 \\[2pt]
       & HyperAgents                      & 265 & 80 & 46.2 & 53.8 & 0.0 \\
\hline
\end{tabularx}
\endgroup
\caption{\textbf{Trajectory comparison for execution-harness refinement.} We compare HyperDSH variants with the main-experiment HyperAgents agent on Combat, reporting recorded and retained opinion counts and the modification-surface distribution of retained opinions. Configurations correspond to Table~\ref{tab:hyperdsh-feasibility}.}
\label{tab:hyperdsh-trajectory}
\end{table}

Despite permission to modify the entire repository, the main-experiment HyperAgents retains no Meta opinions, suggesting a limited tendency to modify its own improver spontaneously.
In contrast, Table~\ref{tab:hyperdsh-trajectory} shows that Meta accounts for 20.5--45.2\% of retained opinions in HyperDSH variants that target DeepSeek Harness refinement and explicitly allow improver self-evolution, reflecting more modifications to the improver itself.
Their retained modifications also shift toward the improver: none of the three variants retains Harness-level opinions, compared with 53.8\% for HyperAgents.
The three HyperDSH variants also produced fewer opinions: 57--127 recorded occurrences, compared with 265 for HyperAgents, indicating less active opinion generation when refining an existing execution harness.
\FloatBarrier




    
    %

\vspace{1em}
\section{Conclusion}
\label{sec:conclusion}

We presented \ourbench, a scalable benchmark for RSI evaluation. Different from conventional benchmarks, \ourbench supports template- and difficulty-controlled task generation for configurable evaluation. We propose a generalization ladder to reflect the transfer to new instances, new difficulties, new tasks within a capability, and across capabilities. We evaluate three representative classes of RSI methods and find that their effectiveness varies across the generalization ladder, with their relative rankings reversing when transferring to tasks of higher difficulty or different types.

\clearpage


\bibliographystyle{IEEEtran}
\bibliography{custom}

\appendix

\section{Evaluation Protocol Details}
\label{sec:evaluation-protocol-details}

\subsection{Task Configurations}
\label{sec:task-configurations}
The task-related fields of $\theta$ determine which instances are generated and how they form $\mathcal{S}_{\mathrm{in}}$ and $\mathcal{S}_{\mathrm{out}}$; model and runtime settings are configured separately.
Game definitions, difficulty parameters, and scoring rules are provided in Appendix~\ref{sec:appendix}.

\paragraph{Held-in tasks.}
The ordered \texttt{schedule} list contains blocks with \texttt{game}, \texttt{difficulty}, and \texttt{count} fields.
Each block generates the requested number of non-replay instances of that game and difficulty.
Blocks are concatenated in list order before replay insertion; \texttt{count} excludes inserted replays.
The block-level \texttt{replay} flag controls eligibility for replay insertion, whose exact semantics are defined in Appendix~\ref{sec:replay-protocol}.
The standard ordered setting uses \texttt{random=false} and \texttt{repeat=false}; these are also the defaults.

\paragraph{Held-out tasks.}
The ordered \texttt{holdout} list specifies evaluation blocks using the same \texttt{game}, \texttt{difficulty}, and \texttt{count} fields.
For each requested game--difficulty pair, \texttt{holdout\_tasks\_per\_difficulty} sets the candidate pool size, while the block's \texttt{count} selects that many instances from the beginning of its pool.
Selected blocks form $\mathcal{S}_{\mathrm{out}}$ in list order, without replay insertion.

\paragraph{Seeds and reconstruction.}
The \texttt{seed} and \texttt{holdout\_seed} fields control held-in and held-out generation, respectively.
For newly generated instances, task seeds are derived deterministically from the root seed, phase, game, difficulty, instance index, and generation attempt using SHA-256.
The current generator rejects duplicate task-payload hashes, excluding \texttt{task\_id}, across the generated held-out pool and non-replay held-in instances; replay insertion deliberately reuses an existing task.
This is descriptor-level deduplication: seed fields remain in the hashed payload, so different descriptors are not guaranteed to produce different underlying game states.
Each run saves its resolved configuration and task manifest as \texttt{config.resolved.json} and \texttt{task\_manifest.json}, including task instances, seeds, sequence order, and replay links.
These artifacts and the corresponding generator version specify the realized sequences.


\subsection{Replay Protocol}
\label{sec:replay-protocol}
\paragraph{Execution.}
Each replay executes the original task instance using the current agent, and records the resulting score for G0 evaluation.
Replay executions do not trigger refinement.

\paragraph{Insertion rule.}
Replays are inserted while constructing the held-in sequence, before execution.
Let $u_1,\ldots,u_n$ be the non-replay tasks placed so far, with $u_n$ belonging to the current schedule block.
After appending $u_n$, insert a replay of $u_{n-\ell}$ if the current block enables replay, the interval $e$ and lag $\ell$ are positive, $n\bmod e=0$, and $n>\ell$.
Both interval and lag count only non-replay tasks, globally across schedule blocks.

\paragraph{Boundary conditions.}
No replay is inserted when history is insufficient ($n\leq\ell$), when the current block disables replay, or when replay is globally disabled ($e=\ell=0$).
No additional replay is inserted to complete a partial interval.
The rule also applies after the final non-replay task: a replay is appended if the same conditions hold.
The ordering above assumes \texttt{random=false}; enabling \texttt{random} shuffles the assembled sequence, including its replay entries, after insertion.

\subsection{Feedback Specification}
\label{sec:feedback-specification}
\paragraph{Abstract interface.}
The benchmark passes task execution results to refinement through a \texttt{FeedbackResult} object,
\[
\mathcal{F}=(\mathrm{id},\mathrm{task},\mathrm{trajectory},\mathrm{evaluation}).
\]
The identifier links feedback to its execution, and \texttt{task} contains the game-specific task descriptor.
The ordered \texttt{trajectory} contains step records with \texttt{step\_index}, \texttt{prompt}, \texttt{image\_records}, \texttt{action}, \texttt{environment\_response}, \texttt{success}, \texttt{score}, \texttt{feedback}, and an optional \texttt{agent\_call}.
The latter retains the model response and, when available, reasoning, token usage, provider, model, and call metadata.
Image records reference available images; text-only interactions use an empty image list.
Step scores, success flags, and feedback may be absent or null when the game does not provide them.

The \texttt{evaluation} object contains execution status, task success, the task score, optional explanatory feedback, and game-specific metrics.
Execution status distinguishes completion, error, timeout, and cancellation; a completed execution need not solve the task successfully.
Metrics depend on the game and can include consumed actions or restart counts.
Game-specific scoring is described in Appendix~\ref{sec:appendix}.

\subsection{Annotation Setup and Validation}
\label{sec:annotation-protocol}
The analysis records use \texttt{gpt-5.6-sol} for semantic annotation.
We checked 100 pre-extracted proposals, stratified by game and method across six games and three methods, against their source modifications.
Table~\ref{tab:annotation-validation} reports agreement with human labels after human-approved, assistant-assisted review with access to machine labels; these are post-review results rather than independent blind agreement estimates.

\begin{table*}[t]
\centering
\begingroup
\definecolor{annotationheader}{RGB}{24,24,24}
\definecolor{annotationgrid}{RGB}{125,125,125}
\definecolor{annotationgroup}{RGB}{245,245,245}
\newcommand{\annotationheading}[1]{\textcolor{white}{\textbf{#1}}}
\setlength{\arrayrulewidth}{0.3pt}
\renewcommand{\arraystretch}{1.15}
\arrayrulecolor{annotationgrid}
\begin{tabularx}{0.6\textwidth}{|>{\columncolor{annotationgroup}\bfseries\raggedright\arraybackslash}X|r|}
\hline
\rowcolor{annotationheader}
\annotationheading{Measure} & \annotationheading{After review} \\
\hline
Scope exact agreement & 88\% \\
\hline
Surface exact agreement & 99\% \\
\hline
Function any-label overlap & 96\% \\
\hline
Function mean Jaccard & 0.828 \\
\hline
\end{tabularx}
\arrayrulecolor{black}
\endgroup
\caption{Agreement between fixed machine labels and reviewed human labels on 100 sampled proposals.}
\label{tab:annotation-validation}
\end{table*}

Scope and Surface each take one label.
Function permits multiple labels because a single proposal can affect several functions, such as reasoning and planning.
Complete-set agreement is therefore stricter than any-label overlap, which requires only $H_i\cap M_i\neq\varnothing$ for human and machine label sets $H_i$ and $M_i$.
Mean Jaccard similarity averages $|H_i\cap M_i|/|H_i\cup M_i|$ across proposals, accounting for both missing and additional labels.
The results indicate strong correspondence after review, particularly for Surface, while differences remain in Scope and complete Function combinations.
The 96\% Function overlap indicates a shared functional label on most proposals, not complete agreement or classification accuracy.


\section{Game Design}
\label{sec:appendix}

In this appendix, we describe the game design in our dataset and detail how game instances are generated by combining task templates with difficulty coefficients.

\subsection{Planning Games}

\subsubsection{Task Definition}

\noindent\textbf{Farm.}
In this game, an agent operates a farm under cash, land, and time constraints. Each task specifies the initial cash, number of plots, planning days horizon, and a task-specific crop catalog containing seed prices, maturation times, harvest yields, and selling prices.  The agent receives the entire instance in one non-interactive call and returns a JSON plan indexed by day. Within each day, actions are executed in the fixed order buy, plant, harvest, sell. The objective is to maximize terminal cash profit, i.e., final cash minus initial cash.

\noindent\textbf{Combat.}
Combat is a deterministic deck-building game comprising alternating combats and shops followed by a boss encounter. Player health, gold, and deck composition persist across steps. At a combat step, the agent observes its current state, owned cards, the enemy's repeating public intent cycle, and the exact deterministic hands. It returns the cards to play on selected turns subject to hand multiplicity and energy constraints. At a shop step, the agent may buy each offered card and purchase one heal, all within its current gold budget. The goal is to defeat the complete encounter chain, including the boss, while maximizing an objective that rewards defeated enemies, remaining health, and remaining gold.

\subsubsection{Task Generation}

\noindent\textbf{Farm.}
Let $d$ denote difficulty.  The generator fixes initial cash to $200$, samples $2d+2$ crops uniformly without replacement from the global catalog, and sets both the number of plots and the days horizon to $4d+4$.  Thus, increasing $d$ simultaneously enlarges the action vocabulary, land capacity, and temporal planning depth. Crops are not tiered, and instance variation arises from the sampled combinations of crop prices, yields, and maturation times.

\noindent\textbf{Combat.}
For difficulty $d\in\{1,2,3,4,5\}$, the available catalog contains every card of tier at most $d$.  The initial deck contains a fixed card set, plus one uniformly sampled card from each tier $2,\ldots,d$. Player parameters remain fixed at $80$ maximum and initial health, $120$ initial gold, $3$ energy, and $5$ drawn cards per turn. Each combat permits at most $15$ turns.  For every tier $i\leq d$, the generator samples one non-boss enemy of tier $i$ as the enemy, and then appends a shop.  Each shop offers three distinct cards. One offer is forced to have tier $\min(d,i+1)$, while the other two are sampled from cards at or below that tier. A tier-$d$ boss is at last, producing $d+1$ combats and $2d+1$ total steps. Enemies and cards become stronger with tier, while the longer step chain increases resource coupling across decisions. 

\subsubsection{References Calculation}

Planning a game involves complex permutations and combinations, and it is difficult to directly compute an optimal solution for it. Therefore, specific search algorithms are used to find a near-optimal solution within a limited time as a pseudo ground-truth.

\noindent\textbf{Farm.}
 The pseudo ground-truth for each farm game is derived from a rolling search. On each day, matured batches are first harvested and sold. Then for each candidate crop, the heuristic tries to greedily plant it into as many free plots as cash permits, and rolls the resulting state to the days horizon using the same daily greedy rule. The candidate with the largest projected terminal cash is committed, and the pseudo ground-truth profit is the terminal cash minus the initial cash. In the above search branches, the daily decision is to plant as much of the same crop as possible, whereas in fact there may be better strategies such as planting a combination of multiple crops. Therefore, the ratio (score) of these strategies compared with the pseudo ground-truth can exceed 1.

\noindent\textbf{Combat.}
Combat computes its reference in two search levels. For each combat stage, the solver tries every legal card combination by turns and simulates its outcome. Among winning plans, the solver prefers higher remaining health and then fewer turns. At each shop, it tries every affordable combination of the three offered cards, both with and without healing. The solver then carries these alternatives through the following stages. To keep the search tractable, partial runs are ranked using remaining health, gold, and the damage and block provided by the deck, and only the top-k plans are continued. After the boss, the completed run with the highest official objective becomes the reference candidate.

\subsection{Reasoning Games}

\subsubsection{Task Definition}

\noindent\textbf{Hero.}
We adapt Hero's Journey~\citep{zheng2026hero}, a text-based adventure benchmark for hidden-rule induction. Given demonstrations and a task description, the agent infers the missing rules and returns a complete action sequence in one response. The objective is to reach the task goal within the action budget.

Each benchmark task contains five separately initialized episodes with different seeds. A successful episode is scored by the ratio of reference solution length to executed action count; an unsuccessful episode receives zero. The task score is the mean episode score.

\noindent\textbf{Sweepmine.}
The agent identifies mines on a partially observed board. Each revealed safe cell indicates the number of mines among its eight neighboring cells. At each step, the agent opens, marks, or unmarks a cell and receives the updated board. The objective is to mark every mine without marking any safe cell within the action budget. Opening a mine generates a new board without resetting the remaining budget.

An unsuccessful task receives zero. For a successful task, the score is $(B-a)/(B-m)$, where $B$ is the action budget, $a$ the number of actions consumed, and $m$ the number of mines.

\begin{table}[h]
\centering
\begingroup
\definecolor{difficultyheader}{RGB}{24,24,24}
\definecolor{difficultygrid}{RGB}{125,125,125}
\definecolor{difficultygroup}{RGB}{245,245,245}
\newcommand{\difficultyheading}[1]{\textcolor{white}{\textbf{#1}}}
\setlength{\arrayrulewidth}{0.3pt}
\renewcommand{\arraystretch}{1.15}
\arrayrulecolor{difficultygrid}
\setlength{\tabcolsep}{3pt}
\begin{tabular}{|>{\columncolor{difficultygroup}\bfseries}l|*{5}{c|}}
\hline
\rowcolor{difficultyheader}
\difficultyheading{Game} & \difficultyheading{d1} & \difficultyheading{d2} & \difficultyheading{d3} & \difficultyheading{d4} & \difficultyheading{d5} \\
\hline
Hero & Ovr/1/3 & Ord/2/2 & Ord/3/2 & Ord/4/2 & Ord/5/2 \\
\hline
Sweepmine & \shortstack{$5\times5$/\\3/16} & \shortstack{$6\times6$/\\5/22} & \shortstack{$7\times7$/\\8/30} & \shortstack{$8\times8$/\\11/40} & \shortstack{$9\times9$/\\15/52} \\
\hline
\end{tabular}
\arrayrulecolor{black}
\endgroup
\caption{Reasoning game difficulty configurations. Hero entries give rule template / hidden action blocks / budget multiplier; Sweepmine entries give board size / mines / action budget.}
\label{tab:reasoning_config}
\end{table}

\subsubsection{Task Generation}

\noindent\textbf{Hero.}
We generate five episodes per task using different seeds, with templates and element pools adapted from the official implementation. Table~\ref{tab:reasoning_config} specifies the rule template, hidden block count, and action-budget multiplier. Override (Ovr) uses class-based defaults with a role-based exception; Ordered (Ord) requires inferring each block's action type, repetition count, and relative order. The action budget is the reference solution length times the multiplier.

\noindent\textbf{Sweepmine.}
We generate boards with seed-controlled mine placement. Difficulty-specific configurations specify the board size as rows $\times$ columns, the total number of mines, and the action budget as the maximum number of attempted actions across all board restarts (Table~\ref{tab:reasoning_config}). The game rules and action interface remain fixed across levels.

\subsection{Exploring Games}
\subsubsection{Task Definition}
\noindent\textbf{MC.}
Minecraft (MC) is a GameWorld browser-based first-person resource-collection and environment-exploration task. The agent starts with an empty inventory and uses visual observations to locate target resources, move, adjust its view, and interact with blocks to complete a collection goal within the step budget. Each step provides the task description, a short interaction history, and the current screenshot, and the agent returns one JSON action.

\noindent\textbf{Astray.}
Astray is a GameWorld maze-navigation task in which the agent uses screenshots to guide a ball to the exit. We adapt the scoring to use the maximum relative reduction in breadth-first search (BFS) shortest-path distance to the exit over the run.

\subsubsection{Task Generation}
\noindent\textbf{MC.}
We define 20 training tasks with different collection goals and target quantities; evaluation uses the five fixed MC tasks from the official GameWorld evaluation set.

\noindent\textbf{Astray.}
Training mazes have side length $L(d)=2d+9$ for difficulty $d\in\{1,\ldots,5\}$, with layouts determined by random seeds. Evaluation uses the five fixed Astray tasks from the official GameWorld evaluation set.

\section{Failure Patterns and RSI Refinement Examples}

Table~\ref{tab:case-refinements} presents failure patterns observed in both baseline systems and initial agent systems of evaluated RSI methods, along with representative refinements proposed by each evaluated RSI method to address these failures.

\begingroup
\definecolor{caseheader}{RGB}{24,24,24}
\definecolor{casegrid}{RGB}{125,125,125}
\definecolor{casegroup}{RGB}{245,245,245}
\definecolor{casecorrect}{RGB}{229,243,231}
\definecolor{caseerror}{RGB}{252,231,231}
\newcommand{\caseheading}[1]{\textcolor{white}{\textbf{#1}}}
\newcommand{\casecomment}[2]{%
  \par\smallskip
  \begingroup
  \setlength{\fboxsep}{2pt}%
  \colorbox{#1}{\parbox{\dimexpr\linewidth-2\fboxsep\relax}{\scriptsize #2}}%
  \endgroup
}
\fontsize{8}{9.4}\selectfont
\setlength{\tabcolsep}{3pt}
\setlength{\arrayrulewidth}{0.3pt}
\renewcommand{\arraystretch}{1.15}
\arrayrulecolor{casegrid}
\setlength{\LTcapwidth}{\linewidth}
\setlength{\LTleft}{0pt}
\setlength{\LTright}{0pt}
\newlength{\casecontentwidth}
\setlength{\casecontentwidth}{\dimexpr\linewidth-12\tabcolsep-7\arrayrulewidth\relax}
\begin{longtable}{|>{\columncolor{casegroup}\bfseries\raggedright\arraybackslash}p{.10\casecontentwidth}|
                   >{\raggedright\arraybackslash}p{.18\casecontentwidth}|
                   *{4}{>{\raggedright\arraybackslash}p{.18\casecontentwidth}|}}
\caption{Failure patterns and example refinements produced by ACE, AHE, HyperAgents,
and HyperDSH. Each pattern was observed in at least one evaluated baseline's traces. 
Red highlights flag flawed assumptions or rules, green highlights flag concrete strategies targeting the stated failure pattern and unhighlighted text presents general refinement guidance.
}
\label{tab:case-refinements}\\
\hline
\rowcolor{caseheader}
\caseheading{Game} & \caseheading{Failure pattern} & \caseheading{ACE refinement} & \caseheading{AHE refinement} & \caseheading{HyperAgents refinement} & \caseheading{HyperDSH refinement} \\
\hline
\endfirsthead
\hline
\rowcolor{caseheader}
\caseheading{Game} & \caseheading{Failure pattern} & \caseheading{ACE refinement} & \caseheading{AHE refinement} & \caseheading{HyperAgents refinement} & \caseheading{HyperDSH refinement} \\
\hline
\endhead
\hline
\endfoot

Hero & Copying demonstrated plans, causing redundant actions.
& \textbf{[Experience]}\newline
Decompose demonstrations and combine only action blocks needed for the target attributes.
& \textbf{[Experience]}\newline
Recombine blocks by target attribute, retaining contributions from unchanged attributes and replacing others only with contrastive evidence.
& \textbf{[Experience]}\newline
Model each feature as \textcolor{red}{an action block} and combine the blocks inferred for the target.
\casecomment{caseerror}{\textcolor{red}{[Error: No guaranteed one-to-one attribute--block mapping.]}}
& \textbf{[Experience]}\newline
Select action blocks by target attributes to build the primary plan and allow \textcolor{green!50!black}{up to two additional rounds} of blocks drawn from other demonstrations.
\casecomment{casecorrect}{\textcolor{green!50!black}{[Correct: Limits retries while allowing alternative action blocks.]}} \\
\hline
Hero & Misinterpreting action types, order, or counts in demonstrations.
& \textbf{[Experience]}\newline
Infer action blocks through controlled comparisons between demonstrations.
& \textbf{[Harness]}\newline
Guide the agent to construct and run shell-based analyses that extract demonstrations and track evidence by attribute.
& \textbf{[Experience]}\newline
Infer action types and counts separately.
& \textbf{[Experience]}\newline
Select action blocks with their counts by target attribute, then check their order and placement against demonstrations. \\
\specialrule{0.6pt}{0pt}{0pt}

Sweepmine & Marking cells incorrectly without checking all current board constraints.
& \textbf{[Experience]}\newline
Check all local constraints across the board and recompute them after each action.
& \textbf{[Experience]}\newline
Use stepwise prompts to rebuild board constraints and incorporate the total mine-count constraint.
& \textbf{[Experience]}\newline
Derive residual constraints from the current board and combine overlapping constraints with the total mine count.
& \textbf{[Experience]}\newline
Use a checklist to check local constraints. A skill also instructs the agent to track feasible mine placements and use the total mine count to guide inference. \\
\specialrule{0.6pt}{0pt}{0pt}

MC & Misidentifying collection targets or whether they can be collected.
& \textbf{[Experience]}\newline
Verify that the target can be mined.
& \textbf{[Harness]}\newline
Verify the target's current state. After a failed mining attempt, \textcolor{green!50!black}{try a different target face}.
\casecomment{casecorrect}{\textcolor{green!50!black}{[Correct: A concrete re-aiming policy after mining failure.]}}
& \textbf{[Experience]}\newline
Verify the target's presence, reachability, and type.
& \textbf{[Experience]}\newline
Check target type and state before collecting, and additionally monitor collection progress. \\
\hline
MC & Failing to adapt when navigation makes no progress.
& \textbf{[Experience]}\newline
Compare position and distance to the resource before and after movement to assess progress, then reorient or excavate if movement is ineffective.
& \textbf{[Harness]}\newline
Switch to excavation toward the resource when exploration stalls.
& \textbf{[Harness]}\newline
\textcolor{green!50!black}{Prevent consecutive repeated actions} and turn or jump when movement stalls.
\casecomment{casecorrect}{\textcolor{green!50!black}{[Correct: Enforces an action-repeat limit in code.]}}
& \textbf{[Experience]}\newline
Compare the current position with the resource's location and stop moving further when the agent is far from it. \\
\specialrule{0.6pt}{0pt}{0pt}

Astray & Misjudging movement direction or traversable paths.
& \textbf{[Experience]}\newline
Assess movement using velocity and displacement along the commanded axis.
& \textbf{[Experience]}\newline
Identify paths from connected floor regions and follow confirmed bends without repeatedly changing direction.
& \textbf{[Experience]}\newline
Assess movement using nearby walls and displacement along the commanded axis.
& \textbf{[Experience]}\newline
Assess movement using structured coordinate changes; use changes in position relative to landmarks instead when coordinate data are unavailable. \\
\hline
Astray & Confusing explored routes and known dead ends with unexplored branches.
& \textbf{[Experience]}\newline
Record exploration history and connectivity to build a map.
& \textbf{[Experience]}\newline
Carry forward a compact MEM ledger of explored branches and confirmed dead ends in each reasoning response.
& \textbf{[Experience]}\newline
Add explicit \textcolor{green!50!black}{wall-following} and backtracking rules.
\casecomment{casecorrect}{\textcolor{green!50!black}{[Correct: A consistent strategy for systematic maze exploration.]}}
& \textbf{[Experience]}\newline
\textcolor{green!50!black}{Follow a wall on a fixed side}, record tried routes and switch to an untried route after consecutive moves away from the goal.
\casecomment{casecorrect}{\textcolor{green!50!black}{[Correct: A consistent strategy for systematic maze exploration.]}} \\
\specialrule{0.6pt}{0pt}{0pt}

Combat & Prioritizing victory or fast kills over remaining HP.
& \textbf{[Experience]}\newline
Compare combat plans by remaining HP.
& \textbf{[Harness]}\newline
Build a \textcolor{green!50!black}{dynamic programming solver} to \textcolor{green!50!black}{maximize remaining HP} after victory.
\casecomment{casecorrect}{\textcolor{green!50!black}{[Correct: Implements HP-oriented search in code.]}}
& \textbf{[Harness]}\newline
Build a \textcolor{green!50!black}{solver to compare complete combat plans} and \textcolor{green!50!black}{maximize remaining HP} after victory.
\casecomment{casecorrect}{\textcolor{green!50!black}{[Correct: Implements HP-oriented search in code.]}}
& \textbf{[Experience]}\newline
Use text instructions to compare combat plans through \textcolor{green!50!black}{dynamic programming} and \textcolor{green!50!black}{maximize remaining HP}.
\casecomment{casecorrect}{\textcolor{green!50!black}{[Correct: Specifies HP-oriented dynamic programming.]}} \\
\end{longtable}
\arrayrulecolor{black}
\endgroup

\clearpage

\end{document}